%% file: main.tex
\documentclass[10pt,letterpaper]{article}
\usepackage[T1]{fontenc}
\usepackage[left=0.72in,right=0.72in,top=0.84in,bottom=0.82in,headheight=23pt,headsep=10pt]{geometry}
\usepackage[table]{xcolor}
\usepackage{newtxtext}
\usepackage{amsmath,amssymb,amsthm}
\usepackage{newtxmath}
\usepackage{graphicx,booktabs,multirow,microtype,url,array,enumitem,float,multicol,placeins}
\usepackage{fancyhdr}
\usepackage{tcolorbox}
\usepackage{titlesec,titletoc}
\usepackage{etoolbox}
\usepackage[colorlinks=true,linkcolor=alilink,citecolor=alilink,urlcolor=alilink]{hyperref}
\hypersetup{
  pdftitle={Calibrating Document-Local Boundaries for Contextual Chunk Retrieval},
  pdfauthor={Yuanxiang Wang, Hui Jiang, Da Zhu, Zhihua Chang}
}

\definecolor{aliorange}{RGB}{255,106,0}
\definecolor{alidark}{RGB}{24,32,44}
\definecolor{alislate}{RGB}{90,101,119}
\definecolor{alilink}{RGB}{18,92,168}
\definecolor{alisurface}{RGB}{249,250,252}
\definecolor{aliheaderbg}{RGB}{241,244,249}
\definecolor{aliborder}{RGB}{216,222,233}
\definecolor{summarybg}{RGB}{247,248,250}
\definecolor{oursbg}{RGB}{242,246,250}
\definecolor{groupbg}{RGB}{248,248,248}

\newcommand{\Lbatch}{\mathcal{L}_{\mathrm{inter}}}
\newcommand{\Lseq}{\mathcal{L}_{\mathrm{seq}}}
\newtheorem{proposition}{Proposition}[section]

\AtBeginEnvironment{thebibliography}{%
  \footnotesize
  \setlength{\itemsep}{0pt plus 0.1pt}
  \setlength{\parskip}{0pt}
  \setlength{\parsep}{0pt}
}

\titleformat{\section}{\large\bfseries\color{alidark}}{\thesection}{0.55em}{}
\titleformat{\subsection}{\normalsize\bfseries\color{alidark}}{\thesubsection}{0.5em}{}
\titleformat{\subsubsection}{\normalsize\bfseries\color{alidark}}{\thesubsubsection}{0.45em}{}
\titleformat{\paragraph}[runin]{\normalsize\bfseries\color{alidark}}{}{0pt}{}[.]
\titlespacing*{\section}{0pt}{10pt plus 2pt minus 2pt}{4pt}
\titlespacing*{\subsection}{0pt}{8pt plus 2pt minus 1pt}{3pt}
\titlespacing*{\subsubsection}{0pt}{7pt plus 1pt minus 1pt}{2.5pt}
\titlespacing*{\paragraph}{0pt}{5pt plus 1pt minus 1pt}{0.6em}

\begin{document}
\twocolumn[{
\begin{@twocolumnfalse}

\noindent
\begin{minipage}{\textwidth}
\setlength{\tabcolsep}{0pt}
\begin{tabular*}{\textwidth}{@{\extracolsep{\fill}}lr@{}}
\raisebox{-0.22\height}{\includegraphics[height=20pt]{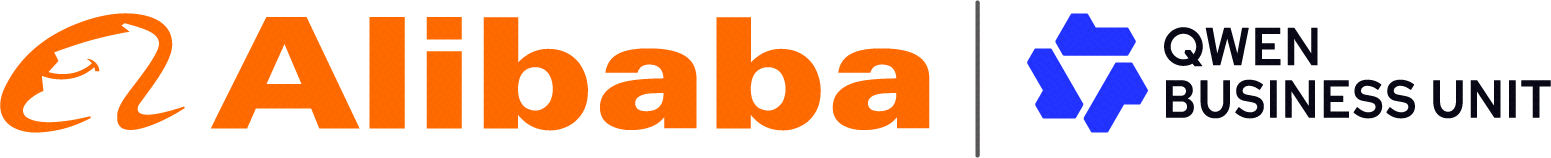}} &
{\footnotesize\bfseries\color{alislate} QWEN BUSINESS UNIT $\cdot$ ALIBABA GROUP}
\end{tabular*}
\vspace{4pt}
{\color{aliorange}\hrule height 1.5pt}
\vspace{14pt}
\begin{center}
{\fontsize{17.5}{22}\selectfont\bfseries\color{alidark} Calibrating Document-Local Boundaries for\\Contextual Chunk Retrieval\par}
\vspace{11pt}
{\normalsize\bfseries Yuanxiang Wang \qquad Hui Jiang \qquad Da Zhu \qquad Zhihua Chang\par}
\vspace{5pt}
{\small\color{alislate} Qwen Business Unit, Alibaba Group\par}
\end{center}
\vspace{6pt}
\end{minipage}

\begin{tcolorbox}[
  colback=alisurface,
  colframe=aliborder,
  boxrule=0.6pt,
  arc=3pt,
  left=14pt,
  right=14pt,
  top=10pt,
  bottom=10pt
]
\noindent{\bfseries\color{alidark} Abstract.}
Late Chunking encodes an entire document before pooling at chunk boundaries, preserving document context in local retrieval units. Chunk-level post-training improves context-dependent retrieval, but shared context can make semantically similar same-document negatives difficult to distinguish.
We study the local decision boundary between a gold chunk and its hardest sibling. Starting from InSeNT, we analyze InSeNT's within-document gradients and show that the entropy of the conditional negative allocation decreases monotonically with the scaling factor at fixed logits.
We propose a two-part framework for calibrating document-local chunk boundaries. The first component adapts the discriminative scale from within-document margins, while the second exposes chunk partitions to attention during document encoding.
Experiments on ConTEB and NanoBEIR show that the resulting model improves contextual chunk retrieval across the evaluated encoder settings while retaining competitive passage-level retrieval quality; ablations further characterize the distinct roles of adaptive scaling and chunk-aware routing.
\par\vspace{5pt}
\noindent{\footnotesize\textbf{\color{alidark} Keywords:} {\color{alislate}Contextual embeddings $\cdot$ Dense retrieval $\cdot$ Contrastive learning $\cdot$ Document representation $\cdot$ Late Chunking}}
\end{tcolorbox}
\vspace{8pt}
\end{@twocolumnfalse}
}]
\thispagestyle{plain}

\section{Introduction}
\label{sec:intro}

Long-document retrieval usually returns local evidence that answers a query, rather than the whole document~\cite{karpukhin2020dpr}. Its meaning often depends on context: pronouns refer to things mentioned earlier, numbers belong to particular years, and similar expressions may describe different facts~\cite{conti2025contextgold}.
Independent chunk encoding loses these cues, whereas a single document vector cannot localize evidence.
Prior work improves contextual retrieval through input enrichment, long-context encoding, and chunk-level post-training~\cite{anthropic2024contextual,gunther2024latechunking,conti2025contextgold}.
These approaches preserve more document context, but retrieval still requires a model to separate locally similar chunks that share that context.

\begin{figure}[t]
\centering
\includegraphics[width=\linewidth]{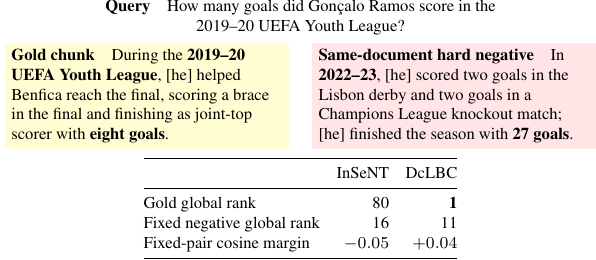}
\caption{Season confusion in Football: the same gold--negative pair under InSeNT and DcLBC.}
\label{fig:badcase}
\end{figure}

Because multiple chunks in a document share the same context, retrieval must distinguish the answer-bearing chunk from its same-document siblings; chunk boundaries therefore directly shape the retrieval objective.
A useful training signal must preserve global context while distinguishing chunks that share entities, topics, or surface form.
Therefore, we treat the gold chunk and its strongest same-document sibling as the basic unit of analysis, and evaluate both contextual and passage-level retrieval.
We focus on contrastive post-training with \emph{same-document negative chunks}.
Of 19,974 InSeNT evaluation queries, 6,976 fail to rank gold first.
Among these 6,976 queries, the highest-scoring sibling exceeds the gold score in 82.21\% of cases, illustrating unresolved local factual confusion.
Figure~\ref{fig:badcase} asks for goals in 2019--20: both passages discuss the same player, European competitions, and scoring records, but the hard negative concerns another season.

We identify two complementary gaps at this local boundary.
First, InSeNT's fixed inverse temperature jointly controls row-wise total effort and its allocation among siblings,
despite heterogeneous query--document margins. AMS instead computes a detached scale from each pair's gold--hardest-sibling margin and applies it only to the InSeNT within-document loss.
Second, distinguishing sibling chunks requires contextual representations that preserve each chunk's local evidence while incorporating shared document context.
Although InSeNT supervises chunks, standard Late Chunking uses partitions only during post-encoder pooling; final-layer attention does not explicitly observe retrieval-unit boundaries.

\begin{figure*}[t]
\centering
\includegraphics[width=0.96\textwidth]{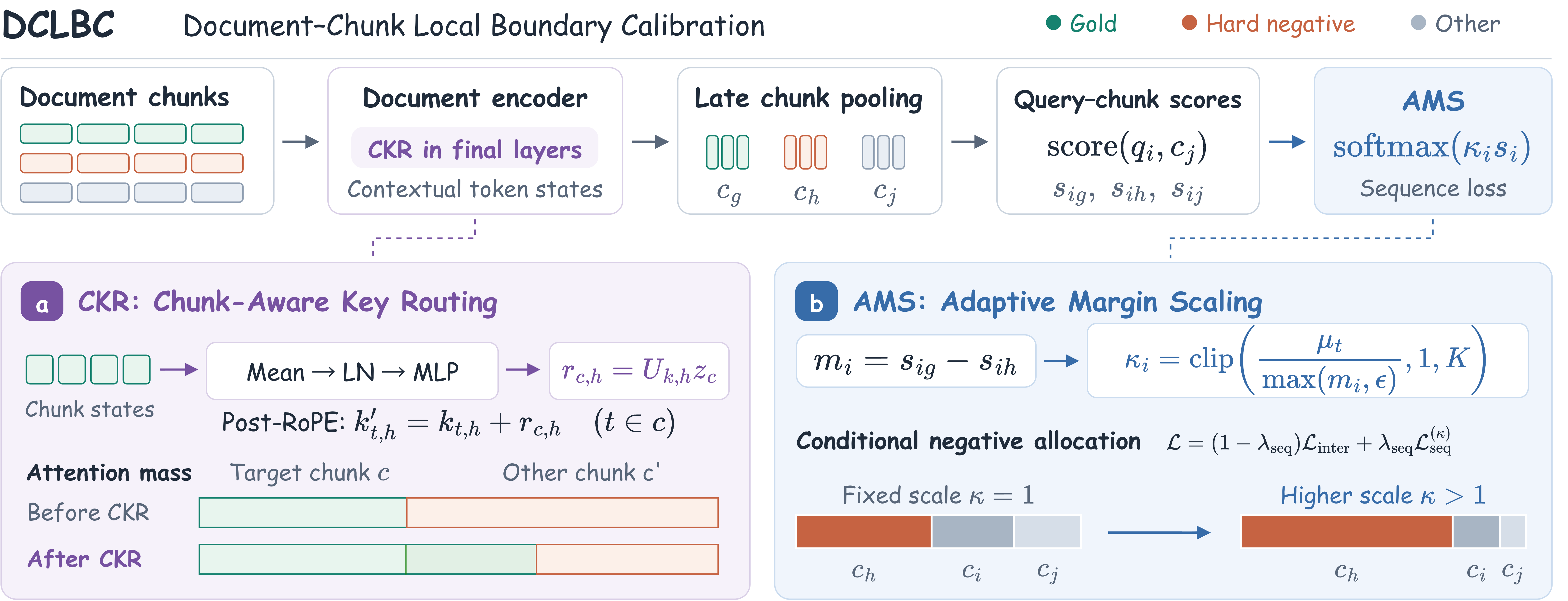}
\caption{Overview of DcLBC. CKR redistributes attention across chunks, while AMS adaptively scales within-document scores.}
\label{fig:method_overview}
\end{figure*}

We propose \textbf{Document-Chunk Local Boundary Calibration (DcLBC)}, combining Adaptive Margin Scaling (AMS) and Chunk-Aware Key Routing (CKR).
AMS adapts the discriminative scale to focus conditional effort on difficult negatives. CKR injects the chunk partition into post-RoPE Keys, allowing chunk-wise context routing during encoding.
Figure~\ref{fig:method_overview} summarizes the pipeline from CKR-enabled encoding through chunk pooling and scoring to the AMS loss. Our contributions are:
\begin{itemize}[leftmargin=1.6em,itemsep=2pt,topsep=3pt]
\item We analyze InSeNT at the query--document row level by separating total score-space effort from its conditional allocation among same-document negatives, and derive how competitor entropy changes with the scale.
\item We introduce DcLBC, combining AMS, which derives a detached instance-level scale from the gold--hardest-sibling margin, with CKR, which injects chunk routing into post-RoPE Keys.
\item We evaluate DcLBC on ConTEB and NanoBEIR with ModernBERT and ModernColBERT backbones, and use component ablations and CKR design tests to measure the contribution of each mechanism.
\end{itemize}

\section{Related Work}
\label{sec:related_work}

\paragraph{Contextualized Chunk Representations.}
Standard passage retrieval encodes each chunk independently, which can discard antecedents, definitions, and other evidence located elsewhere in the source document.
Several approaches therefore enrich retrieval units with broader context.
Contextual Retrieval prepends document-specific explanatory text to each chunk, whereas RAPTOR recursively summarizes and clusters passages into a hierarchical retrieval tree~\cite{anthropic2024contextual,sarthi2024raptor}.
Landmark Embedding learns landmark tokens that organize a long document into semantically retrievable spans rather than relying only on fixed boundaries~\cite{luo2024landmark}.
Contextual Document Embeddings condition an item on related documents, extending contextualization beyond a single document~\cite{morris2025cde}.
Situated Embedding instead trains passage representations to retain information from their surrounding document context~\cite{wu2026situated}.
Late Chunking encodes the full text before pooling token states at chunk boundaries, allowing each pooled chunk to incorporate both left and right context~\cite{gunther2024latechunking}.
Context is Gold further introduces In-Sequence Negative Training, in which chunks from the same document serve as competing candidates during contextual embedding post-training~\cite{conti2025contextgold}.
Together, these studies span textual augmentation, hierarchical organization, learned segmentation, contextual encoding, and chunk-level post-training as complementary routes to context-aware retrieval units.

\paragraph{Long-Context and Retrieval-Oriented Encoders.}
Contextualized retrieval also depends on encoder architectures that can represent long inputs while preserving fine-grained matching signals.
Longformer established an efficient bidirectional design based on sparse local and global attention for processing long sequences~\cite{beltagy2020longformer}, while ModernBERT revisits bidirectional encoder pretraining with modern architectural and training choices~\cite{warner2024modernbert}.
In dense retrieval, DPR represents queries and passages with dual encoders and scores them by vector similarity, and Contriever demonstrates that retrieval representations can also be learned through unsupervised contrastive pretraining~\cite{karpukhin2020dpr,izacard2021contriever}.
Late-interaction models retain token-level vectors: ColBERT performs efficient fine-grained matching with a MaxSim operator, and ColBERTv2 improves its quality and storage efficiency through denoised supervision and residual compression~\cite{khattab2020colbert,santhanam2022colbertv2}.
A parallel line adapts decoder-only language models into embedding encoders.
Echo Embedding repeats the input to expose token representations to preceding context, LLM2Vec combines bidirectional attention with unsupervised and supervised adaptation, and NV-Embed introduces latent attention pooling together with retrieval-oriented training~\cite{springer2025echo,behnamghader2024llm2vec,lee2025nvembed}.
These families provide different trade-offs among context length, representation granularity, scoring expressiveness, and retrieval efficiency.

\paragraph{Contrastive Learning, Hard Negatives, and Temperature.}
Modern embedding models are commonly optimized with contrastive objectives that raise the similarity of matched query--passage pairs relative to negative candidates.
SimCSE shows how contrastive learning can produce strong sentence representations using either dropout-based positive pairs or supervised entailment data~\cite{gao2021simcse}.
For retrieval, negative selection is especially important because randomly sampled passages may become uninformative early in training.
ANCE addresses this issue by periodically mining difficult candidates from an approximate nearest-neighbor index, thereby coupling representation learning with the evolving retrieval space~\cite{xiong2021ance}.
Temperature provides a second control over contrastive optimization: smaller temperatures sharpen the softmax distribution and concentrate gradients on harder negatives~\cite{wang2021behaviour}.
Dual Temperature separates scaling effects associated with different components of contrastive learning to reduce sensitivity to negative-sample quality~\cite{zhang2022dualtemperature}.
MACL further adjusts contrastive training according to model-dependent alignment and uniformity signals rather than using a single fixed optimization regime~\cite{huang2023macl}.
This body of research establishes negative construction and logit scaling as central factors governing the strength and distribution of contrastive learning signals.

\section{Analysis and Method}
\label{sec:method}
\subsection{Within-document chunk competition}
We first recall InSeNT's in-sequence loss~\cite{conti2025contextgold}. Index query--document pairs by $i$ and chunks within pair $i$ by $j$. The normalized query $q_i\in\mathbb{R}^d$ has $n_i$ normalized document chunks $c_{ij}\in\mathbb{R}^d$, collected as $C_i=[c_{i1},\ldots,c_{in_i}]^\top$, with gold index $g_i$. Let $a>0$ be the similarity scale and $\tau>0$ the within-document temperature. Holding the cross-document term fixed,
\begin{equation}
s_{ij}=\frac{a}{\tau}q_i^\top c_{ij},\qquad
\mathcal{L}_{\mathrm{seq}}^{(i)}
=-\log\frac{\exp(s_{ig_i})}{\sum_{j=1}^{n_i}\exp(s_{ij})}.
\label{eq:insent}
\end{equation}
Here $s_{ij}$ is the scaled query--chunk logit. The hardest same-document negative and its margin from gold are
\begin{equation}
\begin{aligned}
h_i&=\arg\max_{j\ne g_i}s_{ij},&
m_i&=s_{ig_i}-s_{ih_i}.
\end{aligned}
\label{eq:margin}
\end{equation}
Smaller $m_i$ means stronger competition. Let $p_{ij}$ be the within-document softmax probability and $\mathbb{I}[\cdot]$ the indicator function:
\begin{equation}
\begin{aligned}
p_{ij}
&=\frac{\exp(s_{ij})}{\sum_{k=1}^{n_i}\exp(s_{ik})},\\
\log\frac{p_{ig_i}}{p_{ih_i}}
&=m_i,\qquad
\frac{\partial\mathcal{L}_{\mathrm{seq}}^{(i)}}{\partial s_{ij}}
 =p_{ij}-\mathbb{I}[j=g_i].
\end{aligned}
\label{eq:margin_coordinate}
\end{equation}
Thus $m_i$ is the gold--hard-negative posterior log-odds around which within-document optimization operates.

\subsection{Document-Chunk Local Boundary Calibration}
Local margins vary across query--document pairs. DcLBC addresses this heterogeneity through AMS and CKR.
\subsubsection{Adaptive Margin Scaling (AMS)}
AMS applies a detached instance-level multiplier $\kappa_i$ to within-document scores, with $p_{ij}^{(\kappa)}=\exp(\kappa_i s_{ij})/\sum_{k=1}^{n_i}\exp(\kappa_i s_{ik})$. The local log-odds become
\begin{equation}
\rho_i(\kappa_i)=\log\frac{p_{ig_i}^{(\kappa)}}{p_{ih_i}^{(\kappa)}}=\kappa_i m_i.
\label{eq:competition_resolution}
\end{equation}

\input{table_main_results}

At training step $t$, let $\mu_t>0$ be the detached EMA reference margin and $\widetilde m_i=\max(m_i,\epsilon)$ the floored margin. Minimizing $(\kappa\widetilde m_i-\mu_t)^2$ over $1\le\kappa\le K$ yields
\begin{equation}
\kappa_i=\operatorname{clip}\left(\frac{\mu_t}{\widetilde m_i},1,K\right).
\label{eq:controller}
\end{equation}
Here $\operatorname{clip}(x,l,u)=\min(\max(x,l),u)$, $K\ge1$ caps the scale, and $\epsilon>0$ prevents divergence for small or negative margins. Appendix~\ref{app:implementation} specifies the EMA update.

\noindent\textbf{Relation to temperature adjustment.}
For untempered scores $u_{ij}=a q_i^\top c_{ij}$, fixed and instance-level temperatures give $s_{ij}=u_{ij}/\tau$ and $u_{ij}/\tau_i$. AMS is also instance-level temperature adjustment, specifically controlled by local same-document competition:
\begin{equation}
\tau_i^{\mathrm{AMS}}=\frac{\tau}{\kappa_i},
\end{equation}
where $\kappa_i$ in Eq.~\eqref{eq:controller} adapts the temperature to each query--document pair's gold--hardest-sibling margin.
This adjustment applies only to the sequence loss; cross-document scores retain their original scale.
Since $\rho_i=\kappa_i m_i$, AMS yields $\rho_i=\mu_t$ whenever $m_i\ge\epsilon$ and $\mu_t/K\le m_i\le\mu_t$.

\noindent\textbf{Optimization pressure.}
Define the total weight and conditional allocation over sibling negatives as
\begin{equation}
W_i=\kappa_i(1-p_{ig_i}^{(\kappa)}),\qquad
r_{ij}=\frac{p_{ij}^{(\kappa)}}{1-p_{ig_i}^{(\kappa)}},
\quad j\ne g_i.
\end{equation}
Let $\mathcal{L}_{\mathrm{DcLBC}}^{(i)}=-\log p_{ig_i}^{(\kappa)}$ be the rescaled row loss, and let $\nabla=\nabla_\theta$ differentiate with respect to trainable parameters $\theta$ while treating $\kappa_i$ as constant. Its negative gradient is
\begin{equation}
-\nabla\mathcal{L}_{\mathrm{DcLBC}}^{(i)}
=W_i\left(\nabla s_{ig_i}-\sum_{j\ne g_i}r_{ij}\nabla s_{ij}\right).
\label{eq:force}
\end{equation}
$W_i$ controls total effort and $r_i$ its allocation over sibling negatives.
At fixed logits, replace $\kappa_i$ in $r_i$ by a generic continuous scale $\kappa>0$ and denote the resulting allocation by $r_i(\kappa)$. Let $H(r_i(\kappa))=-\sum_{j\ne g_i}r_{ij}(\kappa)\log r_{ij}(\kappa)$ be its entropy.
The sibling-only log-partition has second derivative $\operatorname{Var}_{r_i(\kappa)}(s_{ij})$, giving
\begin{equation}
\frac{\mathrm d H(r_i(\kappa))}{\mathrm d\kappa}
=-\kappa\operatorname{Var}_{r_i(\kappa)}(s_{ij})\le0.
\label{eq:entropy}
\end{equation}
At fixed logits, increasing $\kappa_i$ concentrates negative gradients on higher-scoring competitors, providing a more focused update direction for high-competition documents.

\subsubsection{Chunk-Aware Key Routing (CKR)}
CKR introduces chunk boundaries into attention by adding a shared
offset to the keys of each chunk. Consider one attention head
(index omitted), with post-RoPE token queries $q_s\in\mathbb{R}^{d_h}$,
keys $k_t\in\mathbb{R}^{d_h}$, and head dimension $d_h$; $s$ and
$t$ index source and target tokens. From hidden states
$h_t\in\mathbb{R}^{d_{\rm model}}$ in chunk $c$,
we compute a routing vector and update its keys:
\begin{equation}
\begin{aligned}
z_c &= \operatorname{GELU}\!\left(W_{\rm down}
\operatorname{LN}\!\left(\operatorname{Mean}_{t\in c}h_t\right)\right),\\
k'_t &= k_t + U_k z_c,\qquad t\in c.
\end{aligned}
\label{eq:pckr}
\end{equation}
Here $\operatorname{LN}$ denotes layer normalization, $\operatorname{Mean}$ averages token states, and $\operatorname{GELU}$ is the activation. The learned projections $W_{\rm down}\in\mathbb{R}^{r\times d_{\rm model}}$ and $U_k\in\mathbb{R}^{d_h\times r}$ map through routing dimension $r$, so $z_c\in\mathbb{R}^r$.
Adding the offset after RoPE keeps it identical across positions
within a chunk. Thus, for source token $s$, every token in $c$
receives the same additive bias to its attention logit
$\ell_{st}=q_s^\top k_t/\sqrt{d_h}$:
\begin{equation}
\ell'_{st}=\ell_{st}+b_{sc},\qquad
b_{sc}=\frac{q_s^\top U_k z_c}{\sqrt{d_h}}.
\label{eq:ckr_logit_bias}
\end{equation}
Let $a_{st}=\operatorname{softmax}_t(\ell_{st})$ be the original
attention weight and $A_{sc}=\sum_{t\in c}a_{st}$ the total
attention mass assigned to chunk $c$. Applying softmax with
the same mask gives
\begin{equation}
\begin{aligned}
A'_{sc}
&=\frac{e^{b_{sc}}A_{sc}}
{A_{s0}+\sum_{c'}e^{b_{sc'}}A_{sc'}},\\
\frac{a'_{st}}{A'_{sc}}
&=\frac{a_{st}}{A_{sc}},\qquad t\in c,\ A_{sc}>0.
\end{aligned}
\label{eq:chunk_mass}
\end{equation}
Here $A_{s0}$ is the original mass of special or unassigned
tokens, whose keys remain unchanged; all sums respect the
attention mask. CKR therefore multiplies each chunk's mass
by $e^{b_{sc}}$ before renormalization, while preserving its
internal attention proportions for fixed queries, token states,
and routing vectors.

\subsection{Joint training objective}
For a batch of $B$ query--document pairs, DcLBC retains InSeNT's cross-document loss $\Lbatch$ and rescales the within-document loss:
\begin{equation}
\Lseq^{(\kappa)}=-\frac{1}{B}\sum_{i=1}^{B}
\log\frac{\exp(\kappa_i s_{ig_i})}{\sum_{j=1}^{n_i}\exp(\kappa_i s_{ij})}.
\end{equation}
With $\lambda_{\rm seq}\in[0,1]$ denoting the within-document loss weight, the full objective is
\begin{equation}
\mathcal{L}_{\mathrm{DcLBC}}
=(1-\lambda_{\rm seq})\Lbatch+\lambda_{\rm seq}\Lseq^{(\kappa)}.
\label{eq:joint_objective}
\end{equation}
The detached $\kappa_i$ only controls within-document competition. CKR adds no loss: its representations enter both terms, and its parameters are jointly optimized with the encoder.
Complete mathematical proofs of all propositions and identities in this section are provided in Appendix~\ref{app:proofs}.

\section{Experiments}
\label{sec:experiments}
We evaluate the overall retrieval performance of DcLBC and several baseline methods on ConTEB and NanoBEIR, and examine the contributions of AMS and CKR through component ablations and CKR design analyses.
\subsection{Experimental setup}
Following the training data and two-epoch schedule of Context Is Gold~\cite{conti2025contextgold}, all post-training methods use the MLDR, NarrativeQA, and SQuAD corpora. Our runs use 8*A800 GPUs with BF16 and a learning rate of $2\times10^{-4}$, and all methods are evaluated at their final checkpoint. Our primary metric is ConTEB eight-task macro normalized discounted cumulative gain at rank 10 (NDCG@10, abbreviated 8DS); NanoBEIR passage retrieval assesses performance without document context. ModernBERT models use normalized 1024-dimensional cosine scores and ModernColBERT uses MaxSim~\cite{khattab2020colbert,santhanam2022colbertv2}.

\subsection{Main results}
Table~\ref{tab:main_results} compares retrieval models on 8DS and NanoBEIR.
DcLBC consistently outperforms InSeNT on 8DS across the evaluated encoder and multi-vector backbones.
The gains reach 1.40 and 1.91 points on ModernBERT-base and ModernBERT-large, respectively, and peak at 2.45 points on ModernColBERT.
For ModernBERT-large, the largest task-level improvements occur on Football by 6.02 points and Cities by 4.53 points, while COVID is the only contextual task that declines.
This contrast reflects ConTEB's construction: Football and Cities suppress explicit mentions of the document subject to require coreference resolution, whereas COVID-QA often provides strong local matches through extractive medical terminology~\cite{conti2025contextgold}.
This concentration suggests that DcLBC is particularly effective when retrieval depends on distinguishing related chunks within a shared document.

For the two ModernBERT encoders, the NanoBEIR results further show that DcLBC does not trade away retrieval quality when document context is absent. The base and large models retain advantages of 0.44 and 0.81 points over InSeNT, reaching 59.41 and 60.56, respectively. Their concurrent gains on context-dependent ConTEB and passage-level NanoBEIR suggest that local boundary calibration is not merely exploiting extra document context.

\input{table_ablation_design}

\subsection{Ablation and design analysis}
We analyze DcLBC components and CKR design, reporting ConTEB eight-task macro NDCG@10 (\%). Table~\ref{tab:components} first isolates the contributions of AMS and CKR, while Table~\ref{tab:ckr_analysis} examines how the routing configuration affects CKR. Because the two studies use separate controlled comparisons, we interpret each configuration relative to the corresponding baseline in its own table.
\subsubsection{Component ablation}
Table~\ref{tab:components} shows that AMS raises the mean by 1.47 points to 77.59, while CKR alone reaches 76.68, improving over InSeNT by 0.56 points. When added to AMS, CKR contributes a further 0.79-point gain and brings the combined model to 78.38. The joint improvement of 2.26 points is 0.23 points larger than the sum of the two isolated gains, indicating a modest positive interaction rather than simple redundancy. The combined configuration also has the smallest reported standard deviation (0.23) among the four settings. Thus, both components improve retrieval individually, and their combination performs best without increasing the observed across-run variability.

These results are consistent with the distinct roles suggested by our analysis: AMS controls total optimization effort and, at fixed logits, concentrates negative gradient weight on harder siblings, whereas CKR redistributes attention across chunks during encoding. In the population-wide training diagnostics reported in Appendix~\ref{app:effort_routing}, AMS raises the query-row mean calibration effort $\overline W=N^{-1}\sum_{i=1}^{N}W_i$ over the $N$ recorded training rows from 0.374 to 0.535, a 43.1\% increase, further supporting its optimization-side role.

\subsubsection{CKR design analysis}
Table~\ref{tab:ckr_analysis} compares layer coverage and chunk alignment within the AMS-enabled design sweep. Relative to the no-CKR reference at 77.60, routing the last one, two, three, and four selected layers changes the mean to 77.92, 78.05, 77.76, and 78.05, respectively. The non-monotonic pattern shows that increasing the number of routed layers alone does not guarantee a better result. Last~2 and Last~4 attain the same mean, but Last~2 uses fewer routing layers; neither Last~1 nor Last~3 surpasses it. We therefore use Last~2 as the compact default.

Normally, the representation $z_c$ aggregated from chunk $c$ routes the Keys of that same chunk. Cyclically permuting chunk representations within a document breaks this correspondence while retaining the model parameters and other settings. This controlled mismatch reduces the mean from 78.05 to 77.62, a 0.43-point drop. The result indicates that the gain depends not only on adding a routing module, but also on preserving the correspondence between each chunk representation and the chunk whose Keys it modifies.

\subsection{AMS training dynamics}
\begin{figure}[t]
\centering
\includegraphics[width=\linewidth]{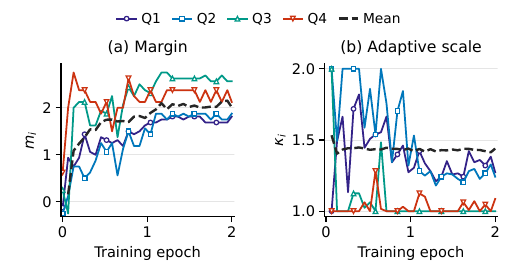}
\caption{Query margins and adaptive scales during DcLBC training. Solid: example queries; dashed: training means.}
\label{fig:calibration_effort}
\end{figure}

We track the gold--hardest-sibling margin $m_i$ and adaptive scale $\kappa_i$ during two epochs of DcLBC training with ModernBERT-large. Figure~\ref{fig:calibration_effort} compares four example queries, measured every ten updates in evaluation mode using the current training EMA, with query-weighted training-batch means.

Q1 and Q2 retain smaller margins and scales above one, whereas Q3's scale settles at one and Q4's fluctuates near this lower bound. This behavior is consistent with Eq.~\eqref{eq:controller}: AMS assigns stronger scaling to queries with smaller margins relative to the evolving reference $\mu_t$. Meanwhile, the mean margin increases, suggesting improved average gold--negative separation, and the mean scale stabilizes near 1.44. These trends are compatible because the reference evolves with training: overall separation can improve while relatively difficult queries retain stronger scaling. These trajectories illustrate AMS's adaptive response; at fixed logits, stronger scaling concentrates negative gradient weight on harder siblings.

\section{Conclusion}
\label{sec:conclusion}
We proposed DcLBC for chunk-level retrieval by combining adaptive margin scaling with chunk-aware context routing. On ConTEB, DcLBC consistently outperforms InSeNT across the evaluated encoder and multi-vector backbones, with gains of up to 2.45 NDCG@10 points. Both ModernBERT variants also improve on NanoBEIR, indicating that stronger contextual retrieval need not compromise passage-level performance. Component ablations and CKR design analyses further support the complementary roles of AMS and CKR in calibrating local chunk boundaries.

\clearpage
\phantomsection
\label{sec:references_start}
\begingroup
\setlength{\parskip}{0pt}
\bibliographystyle{IEEEbib}
\bibliography{references}
\endgroup

\startcontents[appendix]
\input{appendix}
\stopcontents[appendix]

\end{document}

%% file: table_main_results.tex
\begin{table*}[t]
\centering
\caption{ConTEB and NanoBEIR NDCG@10 (\%). Parentheses give base-model parameters in millions (M); bold/underline: best/second best. ``--'' indicates no additional post-training in this study.}
\label{tab:main_results}
\label{tab:per_dataset}
\footnotesize
\setlength{\tabcolsep}{2.5pt}
\renewcommand{\arraystretch}{1.10}
\begin{tabular*}{\textwidth}{@{\extracolsep{\fill}}llrrrrrrrrcc@{}}
\toprule
Model & Post-training
& MLDR & SQuAD & NQA & COVID & Rest. & Foot. & Cities & EIOPA
& \cellcolor{summarybg}\textbf{ConTEB Avg}
& \cellcolor{summarybg}\textbf{NanoBEIR} \\
\midrule
\rowcolor{groupbg}
\multicolumn{12}{@{}l}{\textit{Retrieval baselines without additional post-training}} \\
BM25~\cite{robertson1994okapi} & --
& 69.43 & 56.18 & 74.68 & 53.75 & 19.91 & 12.20 & 45.61 & 0.00
& \cellcolor{summarybg}41.47
& \cellcolor{summarybg}43.41 \\
Contriever-MSMARCO (109M)~\cite{izacard2021contriever} & --
& 75.12 & 70.65 & 77.00 & 53.79 & 42.79 & 24.41 & 67.42 & 14.77
& \cellcolor{summarybg}53.24
& \cellcolor{summarybg}59.54 \\
Anthropic Contextual & --
& 85.40 & 77.10 & 77.70 & 60.70 & 34.80 & 53.90 & 89.40 & \textbf{100.00}
& \cellcolor{summarybg}72.40
& \cellcolor{summarybg}\underline{63.20} \\
\midrule
\rowcolor{groupbg}
\multicolumn{12}{@{}l}{\textit{Encoder models with additional post-training}} \\
& InfoNCE
& 80.09 & 72.62 & 76.84 & 48.54 & 35.48 & 45.84 & 76.47 & \underline{99.69}
& \cellcolor{summarybg}66.95
& \cellcolor{summarybg}58.70 \\
& InSeNT
& 85.81 & 80.11 & 80.33 & 55.81 & 44.66 & 58.25 & 82.92 & \textbf{100.00}
& \cellcolor{summarybg}73.49
& \cellcolor{summarybg}58.97 \\
\multirow{-3}{*}{\shortstack[l]{ModernBERT-base\\(149M)~\cite{warner2024modernbert}}} & \cellcolor{oursbg}DcLBC
& \cellcolor{oursbg}85.28 & \cellcolor{oursbg}80.67 & \cellcolor{oursbg}81.00 & \cellcolor{oursbg}54.59 & \cellcolor{oursbg}44.53 & \cellcolor{oursbg}66.19 & \cellcolor{oursbg}86.83 & \cellcolor{oursbg}\textbf{100.00}
& \cellcolor{oursbg}74.89
& \cellcolor{oursbg}59.41 \\
\cmidrule(lr){1-12}
& InfoNCE
& 81.89 & 72.89 & 78.70 & 51.46 & 44.14 & 63.07 & 85.23 & \textbf{100.00}
& \cellcolor{summarybg}72.17
& \cellcolor{summarybg}59.69 \\
& InSeNT
& 88.92 & \underline{80.80} & \underline{83.51} & 59.53 & 48.25 & 65.49 & 87.10 & \textbf{100.00}
& \cellcolor{summarybg}\underline{76.70}
& \cellcolor{summarybg}59.75 \\
\multirow{-3}{*}{\shortstack[l]{ModernBERT-large\\(395M)~\cite{warner2024modernbert}}} & \cellcolor{oursbg}DcLBC
& \cellcolor{oursbg}89.20 & \cellcolor{oursbg}\textbf{82.92} & \cellcolor{oursbg}\textbf{84.66} & \cellcolor{oursbg}57.80
& \cellcolor{oursbg}\textbf{51.19} & \cellcolor{oursbg}\textbf{71.51} & \cellcolor{oursbg}\underline{91.63} & \cellcolor{oursbg}\textbf{100.00}
& \cellcolor{oursbg}\textbf{78.61}
& \cellcolor{oursbg}60.56 \\
\cmidrule(lr){1-12}
& InfoNCE
& 74.88 & 63.49 & 78.81 & 47.05 & 27.99 & 48.18 & 82.52 & 21.53
& \cellcolor{summarybg}55.56
& \cellcolor{summarybg}54.25 \\
& InSeNT
& \underline{89.32} & 77.24 & 83.32 & \underline{64.78} & \underline{49.00} & 64.55 & 89.66 & 51.42
& \cellcolor{summarybg}71.16
& \cellcolor{summarybg}57.96 \\
\multirow{-3}{*}{\shortstack[l]{ModernColBERT\\(149M)~\cite{chaffin2025gtemoderncolbert}}} & \cellcolor{oursbg}DcLBC
& \cellcolor{oursbg}\textbf{89.70} & \cellcolor{oursbg}74.31 & \cellcolor{oursbg}82.61 & \cellcolor{oursbg}\textbf{65.54} & \cellcolor{oursbg}46.58 & \cellcolor{oursbg}\underline{68.02}
& \cellcolor{oursbg}91.37 & \cellcolor{oursbg}70.73
& \cellcolor{oursbg}73.61
& \cellcolor{oursbg}56.79 \\
\midrule
\rowcolor{groupbg}
\multicolumn{12}{@{}l}{\textit{LLM-based embedding models (causal attention)}} \\
Qwen3-Embedding (596M)~\cite{zhang2025qwen3embedding} & --
& 57.12 & 63.98 & 58.66 & 19.52 & 11.75 & 53.08 & 82.15 & 62.97
& \cellcolor{summarybg}51.15
& \cellcolor{summarybg}\textbf{64.91} \\
Harrier (268M)~\cite{microsoft2026harrier} & --
& 65.77 & 62.30 & 64.36 & 16.83 & 7.71 & 39.30 & 81.74 & 40.20
& \cellcolor{summarybg}47.28
& \cellcolor{summarybg}62.64 \\
ML-Embed (596M)~\cite{codefuse2026mlembed} & --
& 52.25 & 56.21 & 60.19 & 16.15 & 11.14 & 27.91 & 71.54 & 22.00
& \cellcolor{summarybg}39.67
& \cellcolor{summarybg}63.18 \\
\midrule
\rowcolor{groupbg}
\multicolumn{12}{@{}l}{\textit{LLM-based embedding models (bidirectional attention)}} \\
pplx-embed-v1 (596M)~\cite{eslami2026pplxembed} & --
& 84.68 & 74.61 & 81.69 & 53.76 & 26.57 & 62.78 & 82.88 & 86.56
& \cellcolor{summarybg}69.19
& \cellcolor{summarybg}56.07 \\
BidirLM (596M)~\cite{boizard2026bidirlm} & --
& 58.29 & 68.90 & 66.40 & 36.07 & 28.87 & 60.68 & \textbf{92.76} & 23.79
& \cellcolor{summarybg}54.47
& \cellcolor{summarybg}57.45 \\
Nano-Em1-0.6B-v2 (596M)~\cite{kitefish2026nanoem1} & --
& 49.87 & 41.98 & 60.90 & 26.67 & 22.22 & 20.37 & 74.63 & 33.57
& \cellcolor{summarybg}41.28
& \cellcolor{summarybg}49.18 \\
\bottomrule
\end{tabular*}
\end{table*}

%% file: table_ablation_design.tex
\begin{table}[t]
\centering
\caption{Component ablation on ModernBERT-large: ConTEB NDCG@10, mean $\pm$ standard deviation (SD, \%); $\Delta$ in percentage points (pp).}
\label{tab:components}\label{tab:app_factorial}
\small
\renewcommand{\arraystretch}{1.10}
\begin{tabular*}{\linewidth}{@{\extracolsep{\fill}}ccrr@{}}
\toprule
AMS & CKR & ConTEB Avg & $\Delta$ vs.\ InSeNT \\
\midrule
$\times$ & $\times$ & $76.12 \pm 0.59$ & -- \\
$\times$ & $\checkmark$ & $76.68 \pm 0.43$ & $+0.56$ \\
$\checkmark$ & $\times$ & $77.59 \pm 0.33$ & $+1.47$ \\
\rowcolor{oursbg}
$\checkmark$ & $\checkmark$ & $\boldsymbol{78.38 \pm 0.23}$ & $\boldsymbol{+2.26}$ \\
\bottomrule
\end{tabular*}
\end{table}

\begin{table}[t]
\centering
\caption{CKR design analysis on ModernBERT-large: ConTEB NDCG@10, mean $\pm$ standard deviation (SD, \%). Alignment tests use Last~2.}
\label{tab:ckr_analysis}\label{tab:pckr_layers}
\small
\renewcommand{\arraystretch}{1.08}
\begin{tabular*}{\linewidth}{@{\extracolsep{\fill}}lr@{}}
\toprule
Configuration & ConTEB Avg \\
\midrule
\multicolumn{2}{@{}l}{\textit{Layer coverage}} \\
None & $77.60 \pm 0.32$ \\
Last~1 (27) & $77.92 \pm 0.42$ \\
Last~2 (24, 27) & $\boldsymbol{78.05 \pm 0.52}$ \\
Last~3 (21, 24, 27) & $77.76 \pm 0.37$ \\
Last~4 (18, 21, 24, 27) & $78.05 \pm 0.39$ \\
\midrule
\multicolumn{2}{@{}l}{\textit{Chunk alignment}} \\
Correct alignment (Last~2) & $78.05 \pm 0.52$ \\
Cyclic permutation & $77.62 \pm 0.44$ \\
\bottomrule
\end{tabular*}
\par\smallskip
{\footnotesize None disables CKR; other settings share the rank-32 router.}
\end{table}

%% file: appendix.tex
\clearpage
\appendix
\numberwithin{equation}{section}
\numberwithin{table}{section}
\numberwithin{figure}{section}

\twocolumn[{
\begin{@twocolumnfalse}
\begin{center}
{\Large\bfseries Appendix}\par
\vspace{3pt}
{\small\color{alislate} Proofs, implementation details, and supporting analyses}
\end{center}
\vspace{4pt}

\titlecontents{section}[1.6em]
  {\vspace{2pt}\small\bfseries\color{alidark}}
  {\contentslabel{1.6em}}
  {\hspace*{-1.6em}}
  {\titlerule*[0.6pc]{.}\contentspage}
\titlecontents{subsection}[3.8em]
  {\vspace{0.5pt}\footnotesize\color{alidark}}
  {\contentslabel{2.2em}}
  {\hspace*{-2.2em}}
  {\titlerule*[0.6pc]{.}\contentspage}

\begin{tcolorbox}[
  colback=alisurface,
  colframe=aliborder,
  boxrule=0.6pt,
  arc=3pt,
  left=10pt,
  right=10pt,
  top=6pt,
  bottom=6pt,
  title={\small\bfseries\color{alidark} Appendix Contents},
  colbacktitle=aliheaderbg,
  coltitle=alidark,
  fonttitle=\bfseries
]
\begin{multicols}{2}
\setlength{\columnsep}{18pt}
\printcontents[appendix]{l}{1}{\setcounter{tocdepth}{2}}
\end{multicols}
\end{tcolorbox}
\vspace{6pt}
\end{@twocolumnfalse}
}]

\section{Supplementary Proofs}
\label{app:proofs}

This section provides supplementary proofs for Section~\ref{sec:method}.

\subsection{Row-Level Properties of In-Sequence Negative Training}
\label{app:proof_row_relative}

\begin{proposition}[Posterior Log-Odds and Score-Space Gradient of InSeNT; Eq.~\eqref{eq:margin_coordinate}]
\label{prop:eq3_proof}
For normalized query $q_i\in\mathbb{R}^d$ with document chunks $C_i=[c_{i1},\ldots,c_{in_i}]^\top\in\mathbb{R}^{n_i\times d}$, gold index $g_i$, similarity scale $a>0$, and temperature $\tau>0$, let $s_{ij}=(a/\tau)q_i^\top c_{ij}$ and $h_i\in\arg\max_{j\ne g_i}s_{ij}$. Define $m_i=s_{ig_i}-s_{ih_i}$ and $p_{ij}=\exp(s_{ij})/\sum_{k=1}^{n_i}\exp(s_{ik})$. Then:
\begin{equation}
\log\frac{p_{ig_i}}{p_{ih_i}}=m_i,\qquad
\frac{\partial\mathcal{L}_{\mathrm{seq}}^{(i)}}{\partial s_{ij}}
=p_{ij}-\mathbb{I}[j=g_i].
\end{equation}
\end{proposition}

\begin{proof}
Let $Z_i=\sum_{k=1}^{n_i}\exp(s_{ik})$ denote the within-document partition function, so that $p_{ij}=\exp(s_{ij})/Z_i$. Forming the ratio between the gold probability $p_{ig_i}$ and the hardest-sibling probability $p_{ih_i}$, the shared normalizer $Z_i$ cancels:
\begin{equation}
\log\frac{p_{ig_i}}{p_{ih_i}}
=\log\frac{\exp(s_{ig_i})/Z_i}{\exp(s_{ih_i})/Z_i}
=s_{ig_i}-s_{ih_i}=m_i.
\end{equation}
Next, write the sequence loss in Eq.~\eqref{eq:insent} as $\mathcal{L}_{\mathrm{seq}}^{(i)}=-\log p_{ig_i}=-s_{ig_i}+\log Z_i$. Differentiating with respect to each scaled logit $s_{ij}$ yields
\begin{equation}
\begin{aligned}
\frac{\partial\mathcal{L}_{\mathrm{seq}}^{(i)}}{\partial s_{ij}}
&=-\mathbb{I}[j=g_i]
  +\frac{1}{Z_i}\frac{\partial Z_i}{\partial s_{ij}}\\
&=-\mathbb{I}[j=g_i]+\frac{\exp(s_{ij})}{Z_i}\\
&=p_{ij}-\mathbb{I}[j=g_i].
\end{aligned}
\end{equation}
Because $p_{ig_i}/p_{ih_i}=\exp(m_i)$, a smaller margin $m_i$ directly corresponds to weaker posterior discrimination between the gold chunk and its strongest same-document competitor.
\end{proof}

\begin{proposition}[Row Shift Invariance and Zero-Sum Relative Operator]
\label{prop:row_relative}
For a single query $i$, let $\mathcal{L}_{\mathrm{seq}}^{(i)}(s_i)=-s_{ig_i}+\log\sum_{j=1}^{n_i}\exp(s_{ij})$, and let $\mathbf{1}_{n_i}$ be the length-$n_i$ all-ones vector. For any scalar shift $b\in\mathbb{R}$, $\mathcal{L}_{\mathrm{seq}}^{(i)}(s_i+b\mathbf{1}_{n_i})=\mathcal{L}_{\mathrm{seq}}^{(i)}(s_i)$, and the score-space gradient satisfies
\begin{equation}
\sum_{j=1}^{n_i}\frac{\partial\mathcal{L}_{\mathrm{seq}}^{(i)}}{\partial s_{ij}}=0.
\label{eq:app_zerosum}
\end{equation}
Consequently, the sequence loss acts purely on relative differences among same-document chunks and is invariant to document-wide score translations.
\end{proposition}

\begin{proof}
Replacing $s_{ij}$ with $s_{ij}+b$ multiplies both numerator and denominator inside the logarithm by $e^b$, leaving $\mathcal{L}_{\mathrm{seq}}^{(i)}$ unchanged. Summing Proposition~\ref{prop:eq3_proof} over $j\in\{1,\ldots,n_i\}$ gives $\sum_{j=1}^{n_i}(p_{ij}-\mathbb{I}[j=g_i])=\sum_{j=1}^{n_i}p_{ij}-1=0$.
\end{proof}

\subsection{Adaptive Margin Scaling: Scale Calibration, Effort Decomposition, and Entropy Monotonicity}
\label{app:proof_ams}

\begin{proposition}[Constrained Margin-Scale Calibration; Eqs.~\eqref{eq:competition_resolution}--\eqref{eq:controller}]
\label{prop:controller_opt}
Given reference scale $\mu_t>0$, margin floor $\epsilon>0$, and truncated margin $\widetilde{m}_i=\max(m_i,\epsilon)>0$, the box-constrained quadratic calibration problem
\begin{equation}
\kappa_i = \operatorname*{arg\,min}_{\kappa\in[1,K]}\left(\kappa\widetilde{m}_i - \mu_t\right)^2
\end{equation}
admits the unique closed-form solution $\kappa_i = \operatorname{clip}(\mu_t/\widetilde{m}_i, 1, K)$. Furthermore, whenever $m_i\ge\epsilon$ and $\mu_t/K\le m_i\le\mu_t$, the calibrated posterior log-odds between the gold chunk and the hardest sibling satisfies $\rho_i(\kappa_i)=\kappa_i m_i=\mu_t$.
\end{proposition}

\begin{proof}
Because $\widetilde{m}_i>0$, the objective $f(\kappa)=(\widetilde{m}_i)^2(\kappa - \mu_t/\widetilde{m}_i)^2$ is strictly convex in $\kappa$ with unconstrained minimizer $\kappa^\star=\mu_t/\widetilde{m}_i$. Projecting $\kappa^\star$ onto the closed interval $[1,K]$ yields $\kappa_i=\min(\max(\mu_t/\widetilde{m}_i,1),K)=\operatorname{clip}(\mu_t/\widetilde{m}_i,1,K)$. Under scaled logits $\kappa_i s_{ij}$, the posterior log-odds is
\begin{equation}
\rho_i(\kappa_i)=\log\frac{p_{ig_i}^{(\kappa)}}{p_{ih_i}^{(\kappa)}}
=\kappa_i s_{ig_i}-\kappa_i s_{ih_i}=\kappa_i m_i.
\end{equation}
If $m_i\ge\epsilon$ and $\mu_t/K\le m_i\le\mu_t$, then $\widetilde{m}_i=m_i$ and $\mu_t/\widetilde{m}_i\in[1,K]$, so $\kappa_i=\mu_t/m_i$ and $\rho_i(\kappa_i)=\mu_t$. For easy instances ($m_i>\mu_t$), $\kappa_i=1$ preserves the baseline scale; for unresolved or inverted margins ($m_i<\mu_t/K$), $\kappa_i=K$ caps the maximum sharpening multiplier.
\end{proof}

\begin{proposition}[Optimization Effort and Conditional Negative Allocation; Eq.~\eqref{eq:force}]
\label{prop:force_decomposition}
For a detached row multiplier $\kappa_i>0$, let $p_{ij}^{(\kappa)}=\exp(\kappa_i s_{ij})/\sum_{k=1}^{n_i}\exp(\kappa_i s_{ik})$, $W_i=\kappa_i(1-p_{ig_i}^{(\kappa)})$, and $r_{ij}=p_{ij}^{(\kappa)}/(1-p_{ig_i}^{(\kappa)})$ for $j\ne g_i$. Define $\mathcal{L}_{\mathrm{DcLBC}}^{(i)}=-\log p_{ig_i}^{(\kappa)}$ and $\nabla=\nabla_\theta$ with respect to trainable parameters $\theta$, holding $\kappa_i$ fixed. Then $r_i$ is a probability distribution over sibling negatives, and
\begin{equation}
-\nabla\mathcal{L}_{\mathrm{DcLBC}}^{(i)}
=W_i\left(\nabla s_{ig_i}-\sum_{j\ne g_i}r_{ij}\nabla s_{ij}\right).
\end{equation}
\end{proposition}

\begin{proof}
Every $p_{ij}^{(\kappa)}$ is positive, and the probabilities sum to one.
Hence $\sum_{j\ne g_i}p_{ij}^{(\kappa)}=1-p_{ig_i}^{(\kappa)}>0$, which gives $r_{ij}>0$ and $\sum_{j\ne g_i}r_{ij}=1$.
Because $\kappa_i$ is detached in the computation graph, applying the chain rule to $\mathcal{L}_{\mathrm{DcLBC}}^{(i)}=-\kappa_i s_{ig_i}+\log\sum_{j=1}^{n_i}\exp(\kappa_i s_{ij})$ yields
\begin{align}
-\nabla\mathcal{L}_{\mathrm{DcLBC}}^{(i)}
&= -\sum_{j=1}^{n_i}\kappa_i\left(p_{ij}^{(\kappa)}-\mathbb{I}[j=g_i]\right)\nabla s_{ij} \nonumber\\
&= \kappa_i\left(1-p_{ig_i}^{(\kappa)}\right)\nabla s_{ig_i}
   -\sum_{j\ne g_i}\kappa_i p_{ij}^{(\kappa)}\nabla s_{ij} \nonumber\\
&= W_i\nabla s_{ig_i}
   -W_i\sum_{j\ne g_i}
   \frac{p_{ij}^{(\kappa)}}{1-p_{ig_i}^{(\kappa)}}\nabla s_{ij}
   \nonumber\\
&= W_i\left(\nabla s_{ig_i}
   -\sum_{j\ne g_i}r_{ij}\nabla s_{ij}\right).
\end{align}
Thus $W_i$ governs the total magnitude of the score-space update, while $r_i$ governs how repulsive effort is distributed across same-document negative chunks.
\end{proof}

\begin{proposition}[Monotonic Decrease of Conditional Competitor Entropy; Eq.~\eqref{eq:entropy}]
\label{prop:negative_entropy}
At fixed logits $\{s_{ij}\}_{j\ne g_i}$, let $H(r_i)=-\sum_{j\ne g_i}r_{ij}\log r_{ij}$ be the Shannon entropy of the conditional negative allocation $r_i(\kappa)$. Then
\begin{equation}
\frac{\mathrm{d}H(r_i)}{\mathrm{d}\kappa}
=-\kappa\operatorname{Var}_{r_i}(s_{ij})\le 0,
\end{equation}
with strict inequality whenever the sibling negative logits $\{s_{ij}\}_{j\ne g_i}$ are not all identical. Moreover, for any pair of sibling negatives $j,k\ne g_i$ with $s_{ij}>s_{ik}$, the relative weight ratio $r_{ij}/r_{ik}=\exp(\kappa(s_{ij}-s_{ik}))$ grows strictly with $\kappa$.
\end{proposition}

\begin{proof}
Define the negative-only log-partition function $\Phi_i(\kappa)=\log\sum_{j\ne g_i}\exp(\kappa s_{ij})$. Because the gold term cancels when conditioning on $j\ne g_i$, the conditional distribution is an exponential family in canonical parameter $\kappa$:
\begin{equation}
r_{ij}(\kappa)=\frac{\exp(\kappa s_{ij})}{\sum_{k\ne g_i}\exp(\kappa s_{ik})}
=\exp\left(\kappa s_{ij}-\Phi_i(\kappa)\right).
\end{equation}
Standard cumulant-generating properties of exponential families give the first and second derivatives of $\Phi_i(\kappa)$:
\begin{equation}
\begin{aligned}
\Phi'_i(\kappa)
&=\sum_{j\ne g_i}r_{ij}s_{ij}
=\mathbb{E}_{r_i}[s_{ij}],\\
\Phi''_i(\kappa)
&=\sum_{j\ne g_i}r_{ij}
  \left(s_{ij}-\mathbb{E}_{r_i}[s_{ij}]\right)^2\\
&=\operatorname{Var}_{r_i}(s_{ij}).
\end{aligned}
\end{equation}
Substituting $\log r_{ij}=\kappa s_{ij}-\Phi_i(\kappa)$ into the Shannon entropy definition yields the Legendre relation
\begin{equation}
H(r_i)=-\sum_{j\ne g_i}r_{ij}\left(\kappa s_{ij}-\Phi_i(\kappa)\right)
=\Phi_i(\kappa)-\kappa\Phi'_i(\kappa).
\end{equation}
Differentiating with respect to $\kappa$ via the product rule cancels the $\Phi'_i(\kappa)$ terms:
\begin{equation}
\frac{\mathrm{d}H(r_i)}{\mathrm{d}\kappa}
=\Phi'_i(\kappa)-\Phi'_i(\kappa)-\kappa\Phi''_i(\kappa)
=-\kappa\operatorname{Var}_{r_i}(s_{ij})\le 0.
\end{equation}
For $\kappa>0$, $\operatorname{Var}_{r_i}(s_{ij})>0$ whenever the non-gold scores are not identical, establishing strict monotonicity.
\end{proof}

\subsection{Chunk-Aware Key Routing: Exact Attention Factorization}
\label{app:proof_ckr}

\begin{proposition}[Post-RoPE Chunk-Mass Factorization and Within-Chunk Ratio Invariance; Eqs.~\eqref{eq:ckr_logit_bias}--\eqref{eq:chunk_mass}]
\label{prop:pckr_factorization}
Consider a single attention head with post-RoPE query $q_s\in\mathbb{R}^{d_h}$ at source token $s$ and post-RoPE keys $k_t\in\mathbb{R}^{d_h}$ at target tokens $t$. Suppose CKR shifts keys inside each routed chunk $c$ by a chunk-shared offset $k'_t=k_t+U_k z_c$ ($t\in c$), while leaving special or unassigned tokens ($t\in c_0$) unrouted ($k'_t=k_t$). Let $\ell_{st}=q_s^\top k_t/\sqrt{d_h}$, $a_{st}=\operatorname{softmax}_t(\ell_{st})$, and $A_{sc}=\sum_{t\in c}a_{st}$. Then under CKR:
\begin{enumerate}
\item The routed chunk attention mass satisfies
\begin{equation}
A'_{sc}=\frac{e^{b_{sc}}A_{sc}}{A_{s0}+\sum_{c'}e^{b_{sc'}}A_{sc'}},\qquad
b_{sc}=\frac{q_s^\top U_k z_c}{\sqrt{d_h}}.
\end{equation}
\item For any chunk $c$ with $A_{sc}>0$ and any token $t\in c$, the conditional within-chunk token weight is invariant:
\begin{equation}
\begin{aligned}
\frac{a'_{st}}{A'_{sc}}&=\frac{a_{st}}{A_{sc}},\\
\frac{a'_{st}}{a'_{su}}&=\frac{a_{st}}{a_{su}},
\qquad t,u\in c.
\end{aligned}
\end{equation}
\end{enumerate}
\end{proposition}

\begin{proof}
For any target token $t\in c$, linearity of the inner product after RoPE gives
\begin{equation}
\ell'_{st}=\frac{q_s^\top(k_t+U_k z_c)}{\sqrt{d_h}}
=\frac{q_s^\top k_t}{\sqrt{d_h}}+\frac{q_s^\top U_k z_c}{\sqrt{d_h}}
=\ell_{st}+b_{sc}.
\end{equation}
Let $Z_s=\sum_{v}\exp(\ell_{sv})$ be the original row normalizer, where $v$ ranges over valid target tokens. Then $Z_{sc}=\sum_{t\in c}\exp(\ell_{st})=Z_s A_{sc}$ and $Z_{s0}=\sum_{t\in c_0}\exp(\ell_{st})=Z_s A_{s0}$. Under the routed logits $\ell'_{st}$, the unnormalized chunk sum factors as
\begin{equation}
\begin{aligned}
Z'_{sc}
&=\sum_{t\in c}\exp(\ell_{st}+b_{sc})\\
&=e^{b_{sc}}\sum_{t\in c}\exp(\ell_{st})\\
&=e^{b_{sc}}Z_{sc}
=Z_s e^{b_{sc}}A_{sc},
\end{aligned}
\end{equation}
while $Z'_{s0}=Z_{s0}=Z_s A_{s0}$. Dividing $Z'_{sc}$ by the new total normalizer $Z'_s=Z'_{s0}+\sum_{c'}Z'_{sc'}$ and canceling the common positive factor $Z_s$ yields the chunk-mass formula in Claim~1. Finally, for any $t\in c$ with $A_{sc}>0$,
\begin{equation}
\frac{a'_{st}}{A'_{sc}}
=\frac{\exp(\ell'_{st})/Z'_s}{Z'_{sc}/Z'_s}
=\frac{e^{b_{sc}}\exp(\ell_{st})}{e^{b_{sc}}Z_{sc}}
=\frac{\exp(\ell_{st})/Z_s}{Z_{sc}/Z_s}
=\frac{a_{st}}{A_{sc}},
\end{equation}
which establishes Claim~2. Note that applying the translation \emph{after} rotary position embedding is essential: if $U_k z_c$ were added before position-dependent rotation $\mathcal{R}_t$, the effective bias $q_s^\top\mathcal{R}_t(U_k z_c)/\sqrt{d_h}$ would vary with token position $t$, breaking within-chunk ratio invariance.
\end{proof}

\subsection{Boundary-Aligned Geometric Work}
\label{app:proof_boundary_work}

\begin{proposition}[Boundary-Aligned Work Identity and Magnitude Bound]
\label{prop:boundary_work}
Let $C_i\in\mathbb{R}^{n_i\times d}$ contain the normalized chunk embeddings of document $i$, let $q_i\in\mathbb{R}^d$ satisfy $\|q_i\|_2=1$, and let $e_j\in\mathbb{R}^{n_i}$ denote the $j$th standard basis vector. Define $v_i=e_{g_i}-e_{h_i}$. For updated embeddings $C'_i$ and displacement $\Delta C_i=C'_i-C_i$ at fixed $(q_i,g_i,h_i)$, the cosine-margin change $\Delta m_i^{\mathrm{cos}}=(c'_{ig_i}-c'_{ih_i})^\top q_i-(c_{ig_i}-c_{ih_i})^\top q_i$ satisfies
\begin{equation}
\begin{aligned}
\Delta m_i^{\mathrm{cos}}
&=v_i^\top\Delta C_i\,q_i\\
&=\left\langle\Delta C_i,\,v_i q_i^\top\right\rangle_F,\\
\left|\Delta m_i^{\mathrm{cos}}\right|
&\le\sqrt{2}\,\|\Delta C_i\|_F.
\end{aligned}
\label{eq:app_boundary_work}
\end{equation}
\end{proposition}

\begin{proof}
Writing the cosine margin as $m_i^{\mathrm{cos}}=v_i^\top C_i q_i$, linearity yields $\Delta m_i^{\mathrm{cos}}=v_i^\top\Delta C_i q_i=\operatorname{tr}((v_i q_i^\top)^\top\Delta C_i)=\langle\Delta C_i, v_i q_i^\top\rangle_F$. Applying the Frobenius Cauchy--Schwarz inequality gives
\begin{equation}
\begin{aligned}
\left|\Delta m_i^{\mathrm{cos}}\right|
&\le\|\Delta C_i\|_F\,\|v_i q_i^\top\|_F\\
&=\|\Delta C_i\|_F\,\|v_i\|_2\,\|q_i\|_2\\
&=\sqrt{2}\,\|\Delta C_i\|_F,
\end{aligned}
\end{equation}
since $\|e_{g_i}-e_{h_i}\|_2=\sqrt{2}$ for $g_i\ne h_i$ and $\|q_i\|_2=1$. Moreover, because $v_i^\top\mathbf{1}_{n_i}=0$, any common shift $\delta\in\mathbb{R}^d$ applied as $\Delta C_i=\mathbf{1}_{n_i}\delta^\top$ contributes zero to $\Delta m_i^{\mathrm{cos}}$. Thus retrieval improvement depends on signed alignment with the rank-one boundary operator $v_i q_i^\top$ rather than raw representation movement or global cluster expansion.
\end{proof}

\section{Implementation Details}
\label{app:implementation}

All ModernBERT-large variants use the same pre-tokenized training documents, data order, and evaluation pipeline. Documents are capped at 8,192 tokens and encoded once before Late Chunking mean pooling. Post-training runs for two epochs with a global batch of 64 in BF16. We use AdamW with a peak learning rate of $2\times10^{-4}$, cosine decay, a short warmup, and no weight decay. The inter-document and within-document objectives are combined with $\lambda_{\rm seq}=0.1$; the similarity scale is $a=20$, and the within-document temperature is $1.0$ for InSeNT and $0.8$ for AMS-based variants.

AMS uses $K=2$, margin floor $\epsilon=10^{-4}$, and EMA decay $\gamma=0.99$. Let $\mathcal{B}_t$ be the query rows local to the device at the $t$th within-document loss call and
\begin{equation}
\bar m_t=\frac{1}{|\mathcal{B}_t|}\sum_{i\in\mathcal{B}_t}\max(m_i,\epsilon).
\end{equation}
The first call sets $\mu_0=\bar m_0$; subsequent calls update
\begin{equation}
\mu_t=\gamma\mu_{t-1}+(1-\gamma)\bar m_t.
\end{equation}
This EMA is not synchronized across devices, and both $\mu_t$ and $\kappa_i$ are detached from gradient computation. CKR is applied to the final two global-attention layers through one shared rank-32 bottleneck. Its bias-free Key up-projection is zero-initialized, so post-training begins from the pretrained attention distribution.

The resulting chunk vectors are L2-normalized and retrieved with a single cosine score. AMS changes only the training loss and therefore introduces no inference-time operation. CKR adds only the shared bottleneck while leaving the embedding dimension and retrieval procedure unchanged.

\section{Boundary and Optimization Diagnostics}
\label{app:extended_experiments}

\subsection{Competition-Stratified Retrieval Analysis}
\label{app:competition_bins}

\begin{figure*}[t]
\centering
\includegraphics[width=0.74\linewidth]{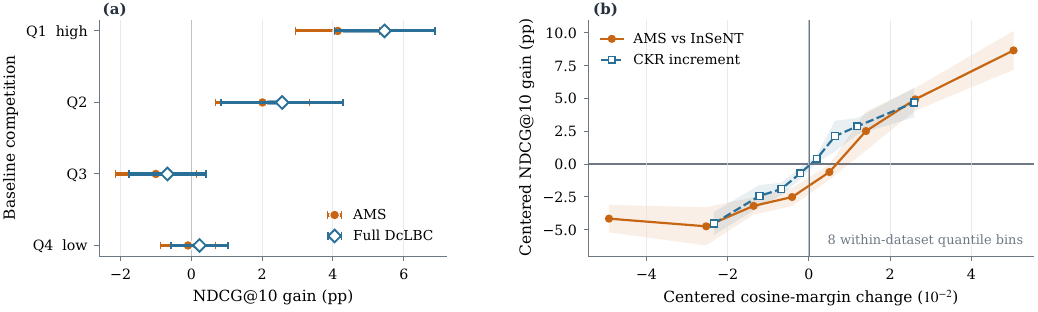}
\caption{\protect\footnotesize Boundary diagnostics. (a) AMS and full DcLBC gains by baseline margin quartile; connectors show the CKR increment and error bars show the dataset-level standard error of the mean (SEM). (b) Within-dataset centered margin and NDCG@10 changes, aggregated into equal-frequency bins.}
\label{fig:app_comp_boundary}
\end{figure*}

We partition queries within each ConTEB dataset into quartiles ordered by the baseline InSeNT gold--hardest-sibling margin. The first quartile contains the most difficult local competitions. Table~\ref{tab:app_bins} gives the compact numerical summary, while Figure~\ref{fig:app_comp_boundary} shows both the concentration of gains in this regime and the query-level association between boundary-margin changes and retrieval changes.

\begin{table}[t]
\centering
\caption{NDCG@10 gains in percentage points (pp) by baseline InSeNT margin quartile. $Q_1$ denotes the highest-competition queries.}
\label{tab:app_bins}
\scriptsize
\setlength{\tabcolsep}{2.5pt}
\renewcommand{\arraystretch}{1.05}
\begin{tabular*}{\columnwidth}{@{\extracolsep{\fill}}lrrrr@{}}
\toprule
\textbf{Comparison} & \textbf{Q1} & \textbf{Q2} & \textbf{Q3} & \textbf{Q4} \\
\midrule
AMS $-$ InSeNT & $+4.14$ & $+2.01$ & $-1.00$ & $-0.09$ \\
CKR increment & $+1.32$ & $+0.56$ & $+0.33$ & $+0.32$ \\
\rowcolor{oursbg}
\textbf{DcLBC $-$ InSeNT} & $\boldsymbol{+5.46}$ & $\boldsymbol{+2.57}$ & $\boldsymbol{-0.67}$ & $\boldsymbol{+0.23}$ \\
\bottomrule
\end{tabular*}
\end{table}

AMS contributes most strongly in $Q_1$ and $Q_2$, where the baseline gold--sibling margin is smallest, rather than uniformly shifting all retrieval scores. CKR adds a positive increment in every quartile, including those for which the AMS-only change is small or negative. Their profiles are therefore complementary: AMS supplies the largest difficult-regime correction, whereas CKR provides a smaller but broader refinement. Figure~\ref{fig:app_comp_boundary} also shows dataset-level uncertainty and the centered association between changes in boundary margin and retrieval quality.

\subsection{Optimization Dynamics}
\label{app:effort_routing}

\begin{figure}[H]
\centering
\includegraphics[width=0.75\columnwidth]{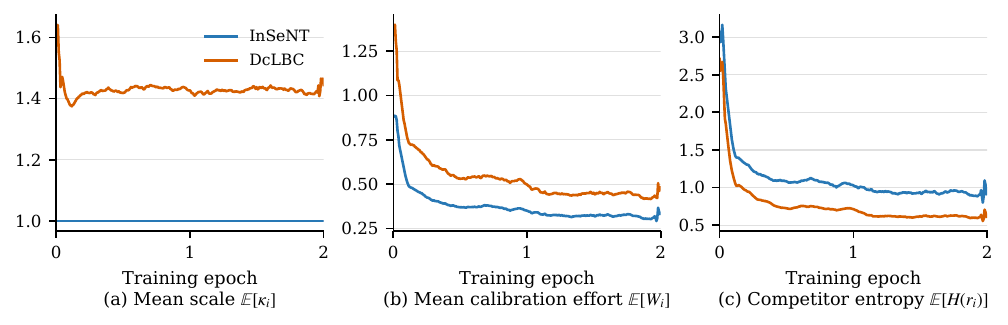}
\caption{\protect\footnotesize Population training dynamics: adaptive scale, calibration effort, and conditional negative-allocation entropy.}
\label{fig:app_opt_diagnostics}
\end{figure}

Figure~\ref{fig:app_opt_diagnostics} shows that, relative to fixed-scale InSeNT, AMS keeps the adaptive scale above one, increases calibration effort, and reduces conditional sibling-allocation entropy. This is consistent with Proposition~\ref{prop:negative_entropy}: difficult rows receive stronger, more concentrated updates. Routed-layer reconstructions further indicate that CKR redistributes attention across chunks and sharpens final-layer target selection.

\FloatBarrier
\section{Robustness and Baseline Comparisons}
\label{app:robustness}

\subsection{Temperature Baseline Comparisons}
\label{app:temperature_baselines}

Table~\ref{tab:app_temperature_full} compares AMS and DcLBC with fixed and adaptive temperature baselines under the same protocol~\cite{zhang2022dualtemperature,huang2023macl}. Controlled-3 is not a separate dataset: it is the macro average over Football, Cities, and EIOPA, the three ConTEB tasks that most directly require document context to resolve a chunk. The comparison shows that global temperature schedules do not match AMS's query-specific document-local calibration.

\begin{table}[t]
\centering
\caption{Temperature baselines on ModernBERT-large (NDCG@10, \%). Controlled-3 is the macro average over the Football, Cities, and EIOPA datasets in ConTEB.}
\label{tab:app_temperature_full}
\footnotesize
\setlength{\tabcolsep}{3pt}
\renewcommand{\arraystretch}{1.06}
\begin{tabular*}{\columnwidth}{@{\extracolsep{\fill}}lrr@{}}
\toprule
\textbf{Method} & \cellcolor{summarybg}\textbf{ConTEB Avg} & \cellcolor{summarybg}\textbf{Controlled-3} \\
\midrule
InSeNT ($\tau=1.0$) & \cellcolor{summarybg}76.70 & \cellcolor{summarybg}84.20 \\
Fixed $\tau=0.8$ & \cellcolor{summarybg}77.15 & \cellcolor{summarybg}85.08 \\
Dual Temperature~\cite{zhang2022dualtemperature} & \cellcolor{summarybg}62.44 & \cellcolor{summarybg}64.28 \\
MACL (temp.-only)~\cite{huang2023macl} & \cellcolor{summarybg}76.15 & \cellcolor{summarybg}82.71 \\
MACL (full)~\cite{huang2023macl} & \cellcolor{summarybg}63.96 & \cellcolor{summarybg}65.84 \\
AMS (Ours) & \cellcolor{summarybg}\underline{77.96} & \cellcolor{summarybg}\underline{87.05} \\
\rowcolor{oursbg}
\textbf{DcLBC (AMS + CKR, Ours)} & \textbf{78.61} & \textbf{87.71} \\
\bottomrule
\end{tabular*}
\end{table}

\section{Additional Encoder Evidence}
\label{app:additional_evidence}

\subsection{Controlled-3 Context-Necessity Subset and Sub-1B LLM Encoders}
\label{app:llm_controlled3}

\begin{table}[t]
\centering
\caption{\protect\footnotesize Aggregate zero-shot comparison of Direct and Late Chunking (LC) modes (NDCG@10, \%). Each score pair is reported as Direct\,/\,LC.}
\label{tab:app_llm_full}
\scriptsize
\setlength{\tabcolsep}{2pt}
\renewcommand{\arraystretch}{0.92}
\begin{tabular*}{\columnwidth}{@{\extracolsep{\fill}}lcc@{}}
\toprule
\textbf{Model} & \textbf{ConTEB} & \textbf{Controlled-3} \\
\midrule
\multicolumn{3}{@{}l}{\textit{Causal encoders}} \\
Qwen3-Embedding~\cite{zhang2025qwen3embedding} & 51.37\,/\,51.15 & 30.85\,/\,66.07 \\
Harrier OSS~\cite{microsoft2026harrier} & 51.54\,/\,47.28 & 35.66\,/\,53.75 \\
ML-Embed~\cite{codefuse2026mlembed} & 45.91\,/\,39.67 & 25.10\,/\,40.48 \\
\midrule
\multicolumn{3}{@{}l}{\textit{Bidirectional encoders}} \\
Giga Embeddings~\cite{aisage2026giga} & 55.75\,/\,63.58 & 35.22\,/\,68.43 \\
pplx-embed-v1~\cite{eslami2026pplxembed} & 46.79\,/\,69.19 & 26.08\,/\,77.41 \\
BidirLM~\cite{boizard2026bidirlm} & 43.02\,/\,54.47 & 23.30\,/\,59.08 \\
Nano-Em1-v2~\cite{kitefish2026nanoem1} & 39.17\,/\,41.28 & 21.81\,/\,42.86 \\
\bottomrule
\end{tabular*}
\end{table}

Controlled-3 is not a standalone dataset but the unweighted macro average of NDCG@10 on Football, Cities, and EIOPA~\cite{conti2025contextgold}. Football and Cities require document-level entity resolution, whereas EIOPA requires disambiguation among structurally similar regulatory sections. This subset focuses on retrieval that requires document context. DcLBC outperforms InSeNT on this subset across all three evaluated backbones.

Table~\ref{tab:app_llm_full} compares independent chunk encoding (\textit{Direct}) with Late Chunking (LC) for zero-shot sub-1B LLM encoders. LC is inconsistent for causal last-token encoders but improves all evaluated bidirectional mean-pooling encoders on both aggregate columns, consistent with the directionality of context propagation.